\documentclass[10pt,twocolumn,letterpaper]{article}

\usepackage{iccv}
\usepackage{times}
\usepackage{epsfig}
\usepackage{graphicx}
\usepackage{amsmath}
\usepackage{amssymb}
\usepackage{graphicx}
\usepackage{amsmath}
\usepackage{amssymb}
\usepackage{booktabs}
\usepackage{multirow}
\usepackage{bm}
\usepackage{enumitem}
\usepackage{arydshln}
\usepackage{adjustbox}
\usepackage{cuted}
\usepackage[accsupp]{axessibility}  

\usepackage[pagebackref=true,breaklinks=true,letterpaper=true,colorlinks,bookmarks=false]{hyperref}

\iccvfinalcopy 

\def\iccvPaperID{3115} 

\ificcvfinal\fi

\newcommand{\nothing}[1]{}

\iccvfinalcopy 

\begin{document}


\title{Mimic3D: Thriving 3D-Aware GANs via 3D-to-2D Imitation}

\author{Xingyu Chen$^*$ \hspace{0.15in}
Yu Deng\thanks{These authors have contributed equally to this work.}
\hspace{0.15in}
Baoyuan Wang \\
Xiaobing.AI \\
}

\maketitle
\ificcvfinal\thispagestyle{empty}\fi

\begin{abstract}
Generating images with both photorealism and multiview 3D consistency is crucial for 3D-aware GANs, yet existing methods struggle to achieve them simultaneously. Improving the photorealism via CNN-based 2D super-resolution can break the strict 3D consistency, while keeping the 3D consistency by learning high-resolution 3D representations for direct rendering often compromises image quality. In this paper, we propose a novel learning strategy, namely 3D-to-2D imitation, which enables a 3D-aware GAN to generate high-quality images while maintaining their strict 3D consistency, by letting the images synthesized by the generator's 3D rendering branch mimic those generated by its 2D super-resolution branch. We also introduce 3D-aware convolutions into the generator for better 3D representation learning, which further improves the image generation quality. With the above strategies, our method reaches FID scores of 5.4 and 4.3 on FFHQ and AFHQ-v2 Cats, respectively, at 512$\times$512 resolution, largely outperforming existing 3D-aware GANs using direct 3D rendering and coming very close to the previous state-of-the-art method that leverages 2D super-resolution. Project website: \url{https://seanchenxy.github.io/Mimic3DWeb}.

\end{abstract}

\section{Introduction}
3D-aware GANs~\cite{schwarz2020graf,gu2021stylenerf,deng2022gram,chan2021efficient} have experienced rapid development in recent years and shown great potential for large-scale realistic 3D content creation. The core of 3D-aware GANs is to incorporate 3D representation learning and differentiable rendering into image-level adversarial learning~\cite{goodfellow2014generative}. In this way, the generated 3D representations are forced to mimic real image distribution from arbitrary viewing angles, resulting in their faithful reconstruction of the underlying 3D structures of the subjects for free-view image synthesis. Among different 3D representations, neural radiance field (NeRF)~\cite{mildenhall2020nerf} has been proven to be effective in the 3D-aware GAN scenario~\cite{schwarz2020graf, chan2021pi}, which guarantees strong 3D consistency when synthesizing multiview images via volume rendering~\cite{kajiya1984ray}.

\begin{figure}[t]
\begin{center}
\includegraphics[width=\linewidth]{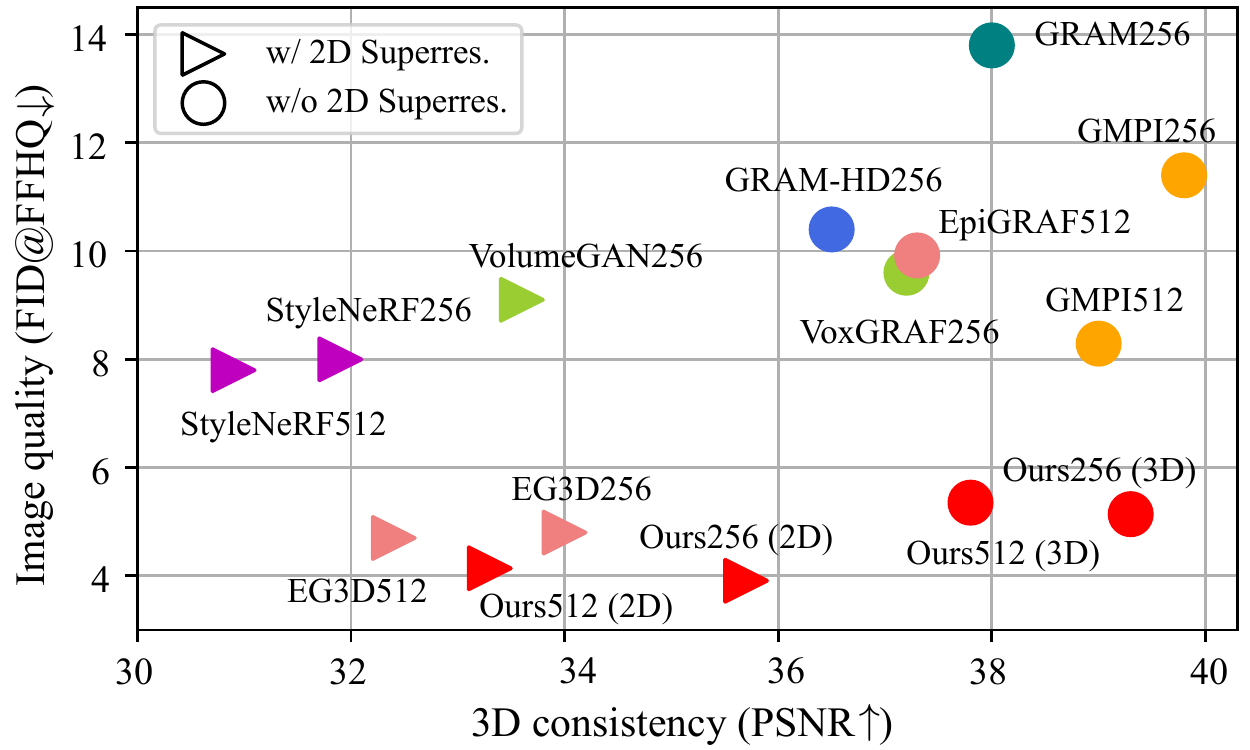}
\caption{Comparison between different 3D-aware GANs on image generation quality and multiview 3D consistency. The image generation quality is evaluated via FID between generated and real images. The 3D consistency is measured by conducting 3D reconstruction~\cite{wang2021neus} on generated multiview images and calculating PSNR between them and the re-rendered reconstruction results. Our method inherits the high image quality of approaches leveraging 2D super-resolution meanwhile maintains strict 3D consistency by taking the advantage of direct 3D rendering. 
}
\label{fig:teaser}
\end{center}
\vspace{-8pt}
\end{figure}

\begin{figure*}[t]
\begin{center}
\includegraphics[width=\linewidth]{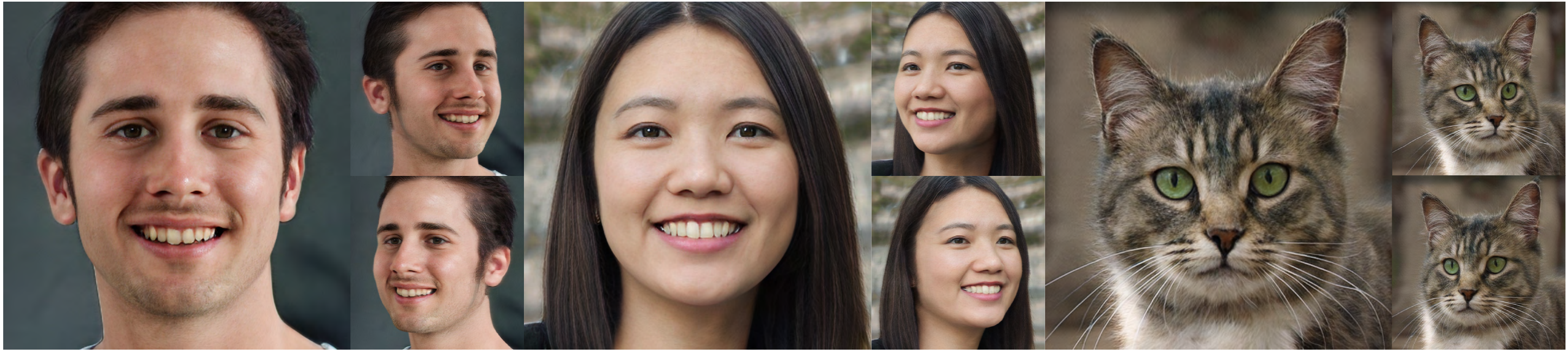}
\caption{Our method enables high-quality image generation at $512\times512$ resolution without using a 2D super-resolution module.}
\label{fig:visual}
\end{center}
\vspace{-8pt}
\end{figure*}

However, NeRF's volumetric representation also brings high computation costs to GAN training. This hinders the generative models from synthesizing high-resolution images with fine details. Several attempts have been made to facilitate NeRF-based GAN training at high resolution, via sparse representations~\cite{schwarz2022voxgraf,deng2022gram,xiang2022gram,zhao2022generative} or patch-wise adversarial learning~\cite{skorokhodov2022epigraf}, yet the performance is still unsatisfactory and lags far behind state-of-the-art 
2D GANs~\cite{karras2020analyzing,karras2021alias}. 

Along another line, instead of using direct NeRF rendering, plenty of works~\cite{niemeyer2021giraffe,gu2021stylenerf,or2022stylesdf,chan2021efficient,xue2022giraffe} introduce 2D super-resolution module to deal with 3D-aware GAN training at high resolution. A typical procedure is to first render a NeRF-like feature field into low-resolution feature maps, then apply a 2D CNN to generate high-resolution images from them. The representative work among this line, namely EG3D~\cite{chan2021efficient}, utilizes tri-plane representation to effectively model the low-resolution feature field and leverages StyleGAN2-like~\cite{karras2020analyzing} super-resolution block to achieve image synthesis at high-quality. It sets a record for image quality among 3D-aware GANs and gets very close to that of state-of-the-art 2D GANs. However, a fatal drawback of this line of works is a sacrifice of strict 3D consistency, due to leveraging a black-box 2D CNN for image synthesis.

A question naturally arises —— \textit{Is there any way to combine the above two lines to achieve strict 3D consistency and high-quality image generation simultaneously?} The answer, as we will show in this paper, is arguably yes. The key intuition is to let the images synthesized by direct NeRF rendering to mimic those generated by a 2D super-resolution module, which we name \textit{3D-to-2D imitation}.

Specifically, we start from an EG3D backbone that adopts 2D super-resolution to generate high-resolution images from a low-resolution feature field. 
Based on this architecture, we add another 3D super-resolution module to generate high-resolution NeRF from the low-resolution feature field and force the images rendered by the former to imitate those generated by the 2D super-resolution branch. This process can be seen as a multiview reconstruction process —— images sharing the same latent code from different views produced by the 2D branch are pseudo multiview data, and the high-resolution NeRF branch represents the 3D scene to be reconstructed. Previous methods~\cite{pan20202d,zhang2020image,poole2022dreamfusion} have shown that this procedure can obtain reasonable 3D reconstruction, even if the multiview data are not strictly 3D consistent. We believe this is partially due to the inductive bias (\eg, continuity and sparsity) of the underlying 3D representation. With the above process, the high-resolution NeRF learns to depict fine details of the 2D-branch images, thus enabling high-quality image rendering. The 3D consistency across different views can also be preserved thanks to the intrinsic property of NeRF. Note that if the rendered images try to faithfully reconstruct every detail of the 2D-branch images across different views, it is likely to obtain blurry results due to detail-level 3D inconsistency of the latter. To avoid this problem, we only let the images produced by the two branches be perceptually similar (\ie by LPIPS loss~\cite{zhang2018unreasonable}), and further enforce adversarial loss between the rendered images from the high-resolution NeRF and real images to maintain high-frequency details. In addition, we only render small image patches to conduct the imitative learning to reduce memory costs.

Apart from the above learning strategy, we introduce 3D-aware convolutions to the EG3D backbone to improve tri-plane learning, motivated by a recent 3D diffusion model~\cite{wang2022rodin}. The original EG3D generates tri-plane features to model the low-resolution feature field via a StyleGAN2-like generator. The generator is forced to learn 2D-unaligned features on the three orthogonal planes via 2D convolutions, which is inefficient. The 3D-aware convolution considers associated features in 3D space when performing 2D convolution, which improves feature communications and helps to produce more reasonable tri-planes. Nevertheless, directly applying 3D-aware convolution in all layers in the generator is unaffordable. As a result, we only apply them after the output layers at each resolution in the tri-plane generator. This helps us to further improve the image generation quality with only a minor increase in the total memory consumption.

With the above strategies, our generator is able to synthesize 3D-consistent images of virtual subjects with high image quality (Fig.~\ref{fig:visual}). It reaches FID scores~\cite{heusel2017gans} of $5.4$ and $4.3$ on FFHQ~\cite{karras2019style} and AFHQ-v2 Cats~\cite{choi2020stargan}, respectively, at $512\times512$ resolution, largely outperforming previous 3D-aware GANs with direct 3D rendering and even surpassing many leveraging 2D super-resolution (Fig.~\ref{fig:teaser}). A by-product of our method is a more powerful 2D-branch generator, which reaches an FID of $4.1$ on FFHQ, exceeding previous state-of-the-art EG3D.
Though our method presented in this paper is mostly based on EG3D backbone, its 3D-to-2D imitation strategy can be extended
to learning other 3D-aware GANs as well. We believe this would largely close the quality gap between 3D-aware GANs and traditional 2D GANs, and pave a new way for realistic 3D generation. 

\section{Related Works}
\paragraph{3D-aware GAN.} 3D-aware GANs~\cite{henzler2019escaping,nguyen2019hologan,schwarz2020graf,chan2021pi,niemeyer2021giraffe,gu2021stylenerf,zhang2022multi,deng2022gram,chan2021efficient,skorokhodov2022epigraf,zhao2022generative} aim to generate multiview images of an object category, given only in-the-wild 2D images as training data. The key is to represent the generated scenes via a 3D representation and leverage corresponding rendering techniques to synthesize images at different viewpoints for image-level adversarial learning~\cite{goodfellow2014generative}. Initially, explicit representations such as voxels~\cite{nguyen2019hologan,henzler2019escaping} and meshes~\cite{szabo2019unsupervised} are used to describe scenes. With the development of neural implicit fields~\cite{park2019deepsdf,mescheder2019occupancy,sitzmann2019scene,sitzmann2020implicit,mildenhall2020nerf,wang2021neus,oechsle2021unisurf}, implicit scene representations, especially NeRF~\cite{mildenhall2020nerf}, gradually overtake explicit ones in 3D-aware GANs~\cite{chan2021pi,or2022stylesdf,chan2021efficient}. Nevertheless, one great hurdle of NeRF-based GANs is the high computation cost, which restricts earlier works~\cite{schwarz2020graf,chan2021pi,devries2021unconstrained,xu2021generative,pan2021shading} from synthesizing high-quality images. Consequently, a large number of follow-up works~\cite{niemeyer2021giraffe,gu2021stylenerf,zhou2021cips,xue2022giraffe,or2022stylesdf,chan2021efficient, xu20223d} avoid rendering NeRF at high resolution by conducting 2D super-resolution from a low-resolution image or feature map rendered by NeRF-like fields. This is only a stopgap as the black-box 2D super-resolution module sacrifices the important 3D consistency brought by NeRF. To keep the strict 3D consistency, several works~\cite{deng2022gram,xiang2022gram,schwarz2022voxgraf,skorokhodov2022epigraf,zhao2022generative} turn to more sparse 3D representations such as sparse voxel~\cite{schwarz2022voxgraf}, radiance manifolds~\cite{deng2022gram}, and multi-plane images~\cite{zhao2022generative} to allow direct rendering at high resolution. Carefully designed training strategies such as two-stage training~\cite{xiang2022gram} or patch-wise optimization~\cite{skorokhodov2022epigraf} are also introduced to facilitate the learning process. However, their image generation quality still lags behind those with 2D super-resolution. Our method combines the advantages of both lines of works to achieve high-quality image generation and strict 3D consistency at once, by leveraging the proposed 3D-to-2D imitation.

\vspace{-8pt}
\paragraph{3D generation by 3D-to-2D imitation.} Recent studies~\cite{jahanian2019steerability, shen2020interpreting,harkonen2020ganspace} reveal that 2D generative models~\cite{brock2018large,karras2019style} have the ability to generate pseudo multiview images of a subject. Based on this observation, several methods~\cite{pan20202d,zhang2020image,shi2021lifting,pan2022gan2x} propose to distill the knowledge from a pre-trained 2D generative model for 3D generation by performing 3D reconstruction on the generated ``multiview" images. A standard procedure is to render the 3D representation of an object from multiple views, and compare them with the closest samples falling in the latent space of the pre-trained 2D generator for iterative optimization. The 2D generator ensures that the rendered results are photorealistic from different views, meanwhile the intrinsic property of the 3D representation guarantees reasonable 3D structure, thus leading to high-quality 3D generation. Some recent methods~\cite{poole2022dreamfusion,lin2022magic3d} also combine this idea with text-to-image diffusion models~\cite{rombach2022high,saharia2022photorealistic} to achieve text-driven 3D creation. Our method shares a similar spirit, which distills the knowledge from the generator's 2D super-resolution branch to its 3D rendering branch, thus achieving image generation with both photorealism and strict 3D consistency.
\begin{figure*}[t]
\begin{center}
\includegraphics[width=0.98\linewidth]{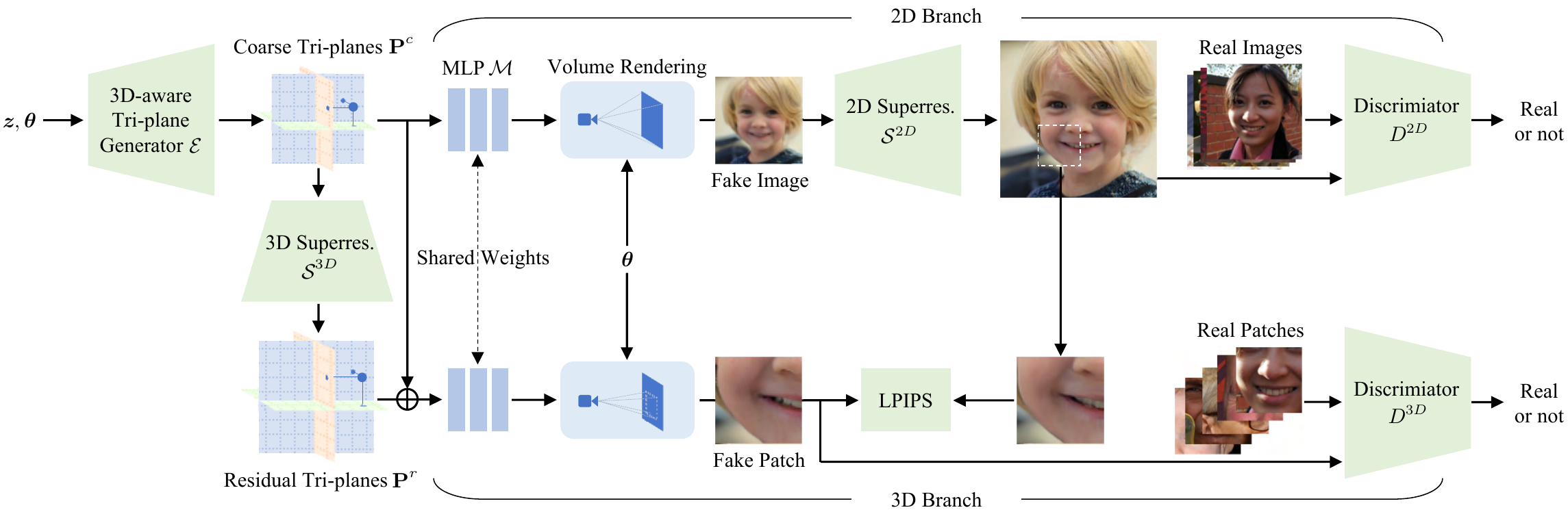}
\caption{Overview of our framework. 3D-to-2D imitation strategy is enforced to let the generator's 3D branch to mimic the results of its 2D branch, thus leading to image generation of high quality and strict 3D consistency. 3D-aware convolutions are also introduced to the tri-plane generator to enhance 3D representation learning, which further improves the image generation quality.}
\label{fig:arch}
\end{center}
\vspace{-0.4cm}
\end{figure*}

\section{Approach}

Given a collection of 2D images, we aim to learn a 3D-aware generator $G$ for free-view image synthesis. The generator takes a random code $\bm z \in \mathbb{R}^{d_z}$ and an explicit camera pose $\bm \theta \in \mathbb{R}^{d_\theta}$ as input, and generates a 2D image $I$:
\begin{equation}
    G: (\bm z,\bm \theta)\in\mathbb{R}^{d_z}\times \mathbb{R}^{d_\theta}\rightarrow I \in \mathbb{R}^{H\times W \times 3}.
\end{equation}
To enable high-quality image synthesis, we adopt EG3D~\cite{chan2021efficient} as the backbone of the generator, which synthesizes low-resolution feature fields via the tri-plane representation~\cite{chan2021efficient}, and leverages 2D super-resolution for high-resolution image generation (Sec.~\ref{sec:eg3d}). Based on EG3D, we propose a 3D-to-2D imitation strategy to synthesize high-resolution NeRF for 3D-consistent image rendering. We leverage a 3D super-resolution branch to predict high-resolution tri-planes from the low-resolution ones, and force the rendered images from the former to mimic the images generated by the 2D super-resolution branch (Sec.~\ref{sec:hr}). In addition, we introduce 3D-aware convolution~\cite{wang2022rodin} to the generator for better tri-plane learning via cross-plane communications, which helps to further improve the image generation quality (Sec.~\ref{sec:aware3d}). The overview of our method is illustrated in Fig.~\ref{fig:arch}. We describe each part in detail below.


\subsection{Preliminaries: EG3D}
\label{sec:eg3d}

EG3D adopts a StyleGAN2-based~\cite{karras2020analyzing} generator $\mathcal E$ to efficiently synthesize the low-resolution feature field of a subject. The feature field is represented by the tri-plane representation which consists of three orthogonal 2D planes produced by reshaping the output feature map of $\mathcal E$, given the latent code $\bm z$ as input. For a point $\bm x \in \mathbb{R}^3$ in the 3D space, its corresponding feature $\mathbf f$ can be obtained by projecting itself onto the three planes $\mathbf{P}_{xy},\mathbf{P}_{yz},\mathbf{P}_{zx}$, and summing the 
retrieved features $\mathbf{f}_{xy},\mathbf{f}_{yz},\mathbf{f}_{zx}$. A small MLP $\mathcal{M}$ then maps this intermediate feature to volume density $\sigma \in \mathbb{R}$ and color feature $\bm c \in \mathbb{R}^{d_c}$ (the first three dimensions represent $RGB$ color), forming the low-resolution feature field:
\begin{equation}
\label{eq:decoder}
\begin{array}{l}
\mathcal{M}: \mathbf{f}\in\mathbb R^{d^{\mathbf f}} \rightarrow (\bm c, \sigma)\in\mathbb R^{d^c}\times\mathbb{R}.
\end{array}
\vspace{-0.1cm}
\end{equation}
To generate high-resolution images, EG3D enforces volume rendering~\cite{kajiya1984ray, mildenhall2020nerf} to render the above feature field to a low-resolution feature map $C$, where each pixel value $C(\bm r)$ corresponding to a viewing ray $\bm r$ can be obtained via
\begin{equation}
\small
\label{eq:render}
	C({\bm r})  
	=   \sum_{i=1}^{N}T_i(1-{\rm exp}(-\sigma_i\delta_i))\bm c_i, T_i={\rm exp}(-\sum_{j=1}^{i-1}\sigma_j\delta_j).
\end{equation}
Here, $i$ is the index of points along ray $\bm r$ sorted from near to far, and $\delta$ is the distance between adjacent points. Then, the rendered feature map $C$ is sent to a 2D super-resolution module $\mathcal S^{2D}$ consisting of several StyleGAN2-modulated convolutional layers to generate the final image~$I^{2D}$.

Although EG3D can generate free-view images of high quality, it cannot well maintain their 3D consistency across different views. This is inevitable due to incorporating the black-box CNN-based 2D super-resolution module, which breaks the physical rules of the volume rendering process. Despite that EG3D further proposes a dual-discrimination~\cite{chan2021efficient} strategy to force the high-resolution images to be consistent with their low-resolution counterparts, detail-level 3D inconsistency (\ie texture flickering) still cannot be eliminated. During continuous camera variation, these artifacts can be easily captured by human eyes, differing the synthesized results from a real video sequence. To maintain the 3D consistency meanwhile keep the high-quality image generation to the maximum extent, we propose a 3D-to-2D imitation strategy described below.


\subsection{3D-to-2D Imitation}
\label{sec:hr}
To keep the strict 3D consistency, a better way is to directly render the 3D representation instead of resorting to a 2D CNN for image synthesis. 
Noticing that the images generated by EG3D contain rich details, it is natural to use them as guidance for images synthesized by direct 3D rendering. If the directly-rendered images well mimic those fine details, their quality should get very close to that of EG3D. Meanwhile, since they are rendered from a continuous 3D representation, their 3D consistency across different views should be trivially maintained. This motivates us to design the 3D-to-2D imitation strategy, as depicted in Fig.~\ref{fig:arch}.

Specifically, we introduce a 3D super-resolution module $\mathcal S^{3D}$ to generate residual tri-planes $\mathbf{P}^r$ from the coarse tri-planes $\mathbf{P}^c$ produced by the tri-plane generator $\mathcal{E}$:
\begin{equation}
\small
    \mathcal S^{3D}: \mathbf{P}^c \in\mathbb{R}^{3\times H^c \times W^c\times d^{\mathbf{f}}}\rightarrow \mathbf{P}^r \in \mathbb{R}^{3\times H^r \times W^r\times d^{\mathbf{f}}}.
\end{equation}
The $\mathcal S^{3D}$ adopts several StyleGAN2-modulated convolutional layers conditioned on a latent code $\bm w$ mapped from the random code $\bm z$, similar to the 2D super-resolution module $\mathcal S^{2D}$ in EG3D. The difference is that $\mathcal S^{3D}$ conducts super-resolution on the triplane-based 3D representation instead of the rendered 2D feature map. In this way, we can generate a high-resolution 3D field for direct 3D rendering. Given the coarse and residual tri-planes (\ie $\mathbf{P}^c$ and $\mathbf{P}^r$), we obtain a more detailed intermediate feature $\mathbf{f} = \mathbf{f}^c + \mathbf{f}^r$ for a 3D point $\bm x$, and further obtain the high-resolution feature field by sending the intermediate feature into the MLP-based decoder $\mathcal{M}$ following Eq.~\eqref{eq:decoder}. The first three feature dimensions of the field derive the high-resolution NeRF for rendering 3D-consistent fine image $I^{3D}$ via Eq.~\eqref{eq:render}.

To ensure that $I^{3D}$ contains reasonable geometry structure with rich texture details, we let it to mimic the contents of $I^{2D}$ generated by the 2D branch $\mathcal{S}^{2D}$. For a pair of $I^{3D}$ and $I^{2D}$ synthesized with the same latent code $\bm z$ and camera pose $\bm \theta$, we enforce imitation loss between them to guarantee their perceptual similarity:
\begin{equation}
\label{eq:imitation}
\begin{array}{l}
\mathcal L_{imitation}=\mathrm{LPIPS}(I^{3D}, \mathrm{sg}(I^{2D})),
\end{array}
\vspace{-0.1cm}
\end{equation}
where $\mathrm{LPIPS}(\cdot,\cdot)$ is the perceptual loss defined in~\cite{zhang2018unreasonable}, and $\mathrm{sg}$ denotes stopping gradient to avoid undesired influence of $I^{3D}$ on the 2D branch. This process is very similar to a standard multiview reconstruction process. During training, $I^{2D}$ sharing the same code $\bm z$ are generated under different camera views from a statistical aspect, forming the multiview supervision. The high-resolution NeRF from the 3D branch renders $I^{3D}$ under the same camera views to compare with the multiview data for 3D reconstruction. Considering that $I^{2D}$ are nearly 3D-consistent, they should help to learn a reasonable NeRF for 3D-consistent image rendering. 

Nevertheless, since $I^{2D}$ are not strictly 3D-consistent, faithfully reconstructing their image contents leads to blurry results where the texture details across different views are averaged out. Therefore, we further introduce the non-saturating GAN loss with R1 regularization~\cite{mescheder2018training} between $I^{3D}$ and real images $\hat{I}$ to maintain the high-frequency details:
\begin{equation}
\label{eq:adv_3d}
\begin{aligned}
\mathcal{L}_{adv}^{3D}& = \mathbb{E}_{\boldsymbol{z} \sim p_z, \boldsymbol{\theta} \sim p_\theta}[f(D^{3D}(G^{3D}(\boldsymbol{z}, \boldsymbol{\theta})))] \\
& +\mathbb{E}_{\hat{I} \sim p_{real}}[f(-D^{3D}(\hat{I}))+\lambda\|\nabla D^{3D}(\hat{I}))\|^2],
\end{aligned}
\vspace{-0.1cm}
\end{equation}
where $f(u)=\log(1+\exp{(u)})$ is the Softplus function, $G^{3D}$ including $\{\mathcal{E},\mathcal{M},\mathcal{S}^{3D}\}$ is the 3D rendering branch of the generator, and $D^{3D}$ is the corresponding discriminator.

An advantage of the above imitation learning is that we can render small patches (\ie $64\times64$) to compute Eq.~\eqref{eq:imitation} and Eq.~\eqref{eq:adv_3d}, as shown in Fig.~\ref{fig:arch}, with only minor influence to the final image quality. 
This largely reduces the memory cost during training and enables learning the 3D branch at high resolution (\eg $512\times512$). By contrast, solely applying adversarial loss at patch-level often leads to large quality drops as shown in previous methods~\cite{schwarz2020graf, skorokhodov2022epigraf} and Tab.~\ref{tab:sr}.


Finally, we apply image-level adversarial loss to the 2D branch following EG3D to ensure that $I^{2D}$, as the supervision for the 3D branch, are of high quality:

\begin{equation}
\label{eq:adv_2d}
\begin{aligned}
\mathcal{L}_{adv}^{2D}& = \mathbb{E}_{\boldsymbol{z} \sim p_z, \boldsymbol{\theta} \sim p_\theta}[f(D^{2D}(G^{2D}(\boldsymbol{z}, \boldsymbol{\theta})))] \\
& +\mathbb{E}_{I \sim p_{real}}[f(-D^{2D}(\hat{I}))+\lambda\|\nabla D^{2D}(\hat{I}))\|^2],
\end{aligned}
\vspace{-0.1cm}
\end{equation}
where $G^{2D}$ is the 2D branch generator consisting of $\{\mathcal{E},\mathcal{M},\mathcal{S}^{2D}\}$, and $D^{2D}$ is the corresponding discriminator. The same dual discrimination is adopted as done in EG3D. 

Overall, the training objective is 
\begin{equation}
\label{eq:loss}
\mathcal L_{total}=\mathcal{L}_{imitation} + \mathcal{L}_{adv}^{3D}+\mathcal{L}_{adv}^{2D}.
\vspace{-0.1cm}
\end{equation}
In practice, we first learn the 2D branch via $\mathcal{L}_{adv}^{2D}$ to obtain reasonable synthesized images $I^{2D}$, then leverage $\mathcal L_{total}$ to simultaneously learn the 2D and 3D branches for high-quality and 3D-consistent image synthesis.

\subsection{3D-Aware Tri-plane Generator}
\label{sec:aware3d}
\begin{figure}[t]
\begin{center}
\includegraphics[width=\linewidth]{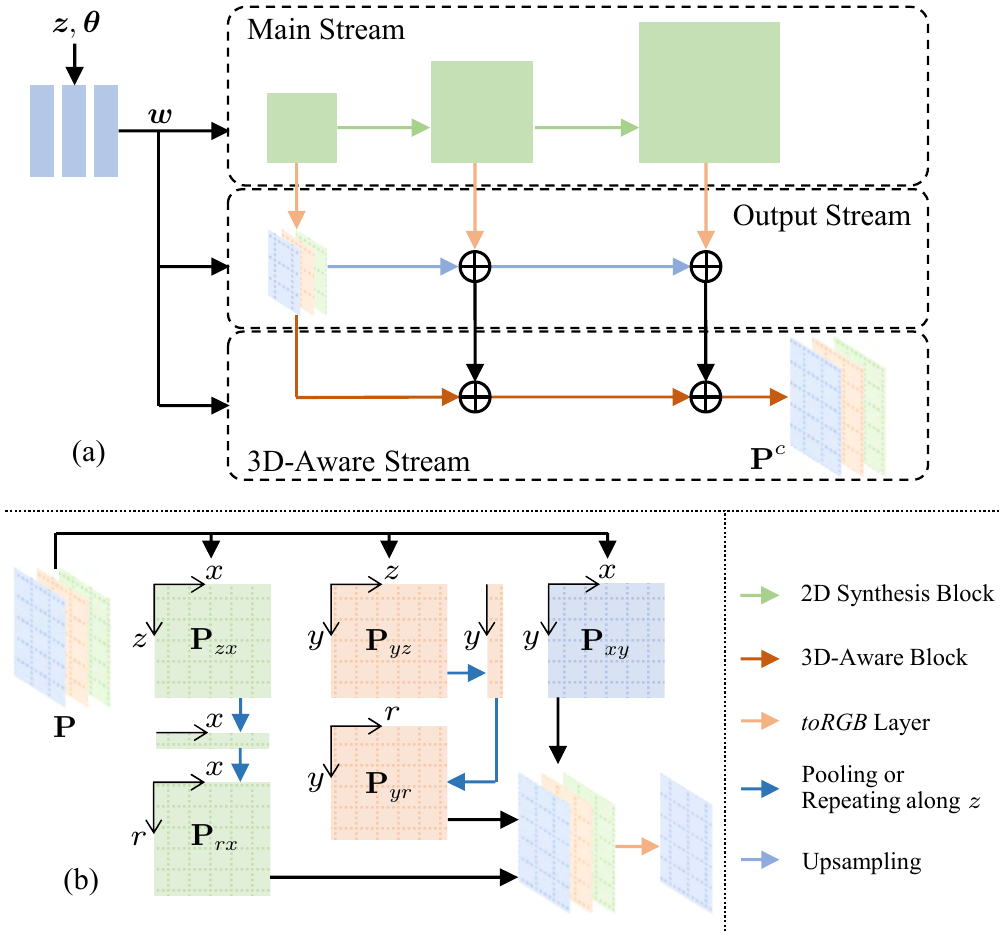}
\caption{(a) Structure of our 3D-aware tri-plane generator. (b) Operations of the 3D-aware block on $xy$ plane.}
\label{fig:araware3d}
\end{center}
\vspace{-0.3cm}
\end{figure}

As depicted in Sec.~\ref{sec:hr}, the tri-plane generator $\mathcal{E}$ is responsible for synthesizing the coarse tri-planes $\mathbf{P}^c$ shared by both the 2D and 3D branches, which is an important component that would affect the final image generation quality. However, in EG3D, $\mathcal{E}$ takes a StyleGAN2 architecture originally designed for 2D generative tasks. As shown in Fig.~\ref{fig:araware3d}(a), the original tri-plane generator only contains the main stream and the output stream. The tri-planes are obtained from latent feature maps in the main stream via 2D convolutions (\ie $toRGB$ layers), and the latent feature maps are also produced by a serials of 2D synthesis blocks. Consequently, the latent feature maps are forced to learn 3D unaligned features of the three orthogonal planes and the latters also lack feature communications with each other. Inspired by a recent 3D diffusion model~\cite{wang2022rodin}, we introduce 3D-aware convolutions into our tri-plane generator $\mathcal{E}$ to enhance feature communications between 3D-associated positions across different planes, for better tri-plane generation. 

\begin{figure*}[t]
\begin{center}
\includegraphics[width=\textwidth]{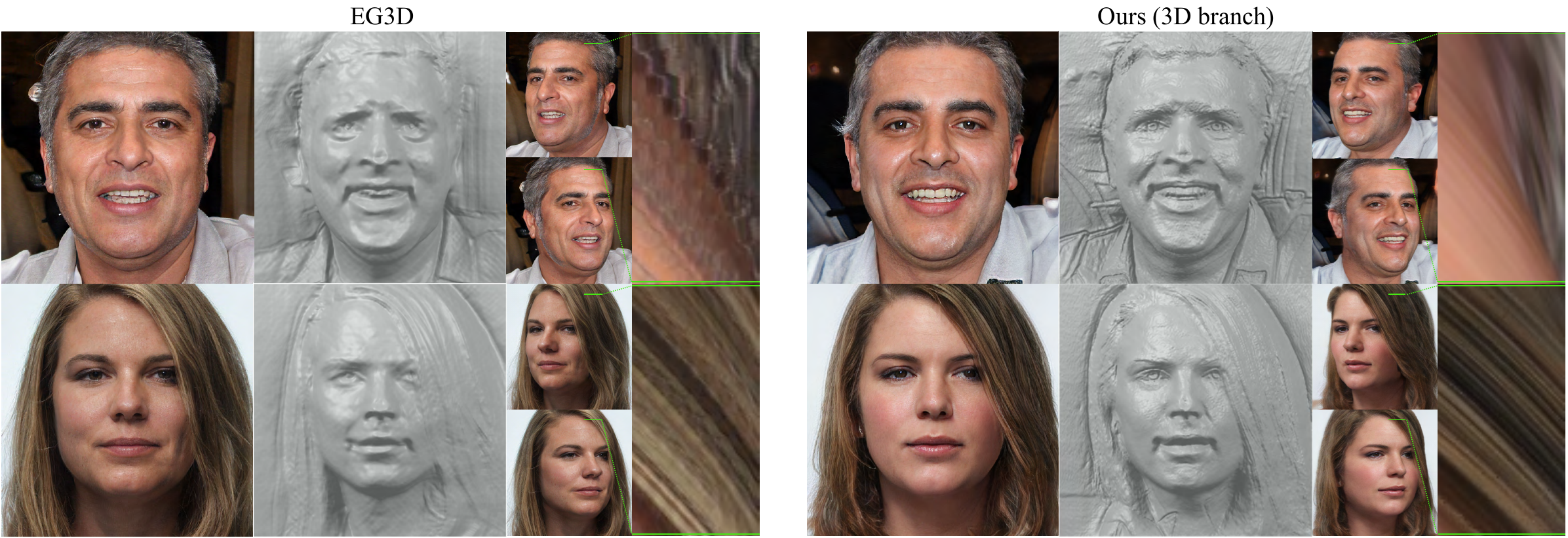}
\caption{Comparison between our method and EG3D on FFHQ at $512\times512$ resolution. Our method generates images with comparable quality to those of EG3D, while producing 3D geometries with finer details and multiview sequences with better 3D-consistency.}
\label{fig:geo}
\end{center}
\vspace{-0.3cm}
\end{figure*}

\begin{figure*}[t]
\begin{center}
\includegraphics[width=\linewidth]{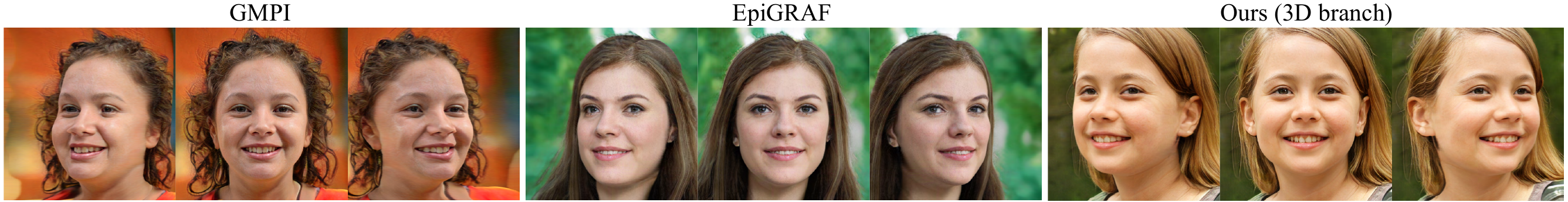}
\caption{Comparison between our method and other 3D rendering baselines on FFHQ at $512\times512$ resolution. \textbf{Best viewed with zoom-in.}}
\label{fig:compare}
\end{center}
\vspace{-0.3cm}
\end{figure*}

Specifically, as illustrated in Fig.~\ref{fig:araware3d}(a), we add an extra 3D-aware stream upon the original output stream after each $toRGB$ layer at different resolutions. At each resolution level $k$, the corresponding tri-planes $\mathbf P_k=[\mathbf P_{k,xy},\mathbf P_{k,yz},\mathbf P_{k,zx}]$, are summed with the tri-planes produced by the original output stream, and further sent into a 3D-aware block to produce tri-plane features for the next level. The 3D-aware block conducts similar operations on each of the three planes. For brevity, we omit the subscript $k$ here and take $\mathbf P_{xy}$ as an example to illustrate the operation process. As shown in Fig.~\ref{fig:araware3d}(b), to align $\mathbf P_{yz}$ and $\mathbf P_{zx}$ towards $\mathbf P_{xy}$, we first perform global pooling along $z$ axis of the former two to obtain $z$-squeezed feature vectors. These vectors are then repeated along the $z$ dimension to restore the original spatial size, denoted as $\mathbf P_{yr}$ and $\mathbf P_{rx}$. In this manner, the obtained $\mathbf P_{yr}$ and $\mathbf P_{rx}$ are aligned with $\mathbf P_{xy}$ from a 3D perspective, \ie, a 2D position $uv$ on $\mathbf P_{xy}$ is responsible for features in region $uvz, z \in [z_{min},z_{max}]$ in the 3D space, meanwhile the same $uv$ position on $\mathbf P_{yr}$ and $\mathbf P_{rx}$ also associate with the features in this 3D region. As a result, we can simply concatenate them along the channel dimension as $[\mathbf P_{xy},\mathbf P_{yr},\mathbf P_{rx}]$, and perform modulated 2D convolution~\cite{karras2020analyzing} on it. The 2D convolution aggregates the 3D-associated features to produce next-level $\mathbf P_{xy}$, leading to better feature communications across the planes. 
$\mathbf P_{yz}$ and $\mathbf P_{zx}$ can be processed similarly.

Note that in~\cite{wang2022rodin}, the 3D-aware convolution is applied in all layers in a U-Net structure. However, in our scenario, leveraging 3D-aware convolution for all layers, especially the main stream, introduces unaffordable memory cost during training, as it would produce multiple auxiliary tensors and triples the channel dimension for each processed latent feature map, as shown in Fig.~\ref{fig:araware3d}(b). Comparing to the latent feature maps in the main stream, the tri-planes after each output layer contains much fewer channels thus more memory-friendly to adopt the 3D-aware convolution. Empirically, our proposed 3D-aware stream helps to learn more reasonable tri-planes and improves the final image generation quality, with only a minor increase in the total memory consumption (see Sec.~\ref{sec:ablation}).

\section{Experiments}

\begin{table*}[t]
\small
\centering
\begin{tabular}{c c | c c c | c c c | c | c  }
\hline
\multicolumn{2}{c|}{\multirow{2}{*}{Method}} & \multicolumn{3}{c|}{FFHQ256} & \multicolumn{3}{c|}{FFHQ512} & CATS256 & CATS512 \\ 
 & &\!\! FID $\downarrow$ \!\!& \!\!PSNR$_{mv}$ $\uparrow$\!\! &\!\! SSIM$_{mv}$ $\uparrow$\!\! & \!\!FID $\downarrow$\!\! & \!\!PSNR$_{mv}$ $\uparrow$ \!\!& \!\!SSIM$_{mv}$ $\uparrow$\!\! & \!\!FID $\downarrow$\!\! & \!\!FID $\downarrow$\!\! \\
\hline 
\multirow{4}{*}{\rotatebox{90}{\textit{w/ 2D SR}}}
& StyleSDF \cite{or2022stylesdf}  & 11.5 & - & - & 11.2 & - & - & - & 7.91 \\
& VolumeGAN \cite{xu20223d}   & 9.10  & 33.6 & 0.926 & -  & - & - & - & - \\
& StyleNeRF \cite{gu2021stylenerf}   & 8.00  & 31.9 & 0.915 & 7.80  & 30.9 & 0.843 & - & - \\
& EG3D \cite{chan2021efficient}      & 4.80  & 34.0 & 0.928 & 4.70  & 32.4 & 0.861 & 3.88 & 2.77 \\
& Ours (2D branch)       & \bf 3.91  & \bf 35.7 & \bf 0.938 & \bf 4.14  & \bf 33.3 & \bf 0.891 & \bf 3.41 & \bf 2.72  \\
\hline
\multirow{6}{*}{\rotatebox{90}{\textit{w/o 2D SR}}}
& GRAM \cite{deng2022gram}              & 13.8 & 38.0& 0.966 & - & - & - & 13.4 & - \\
& GRAM-HD \cite{xiang2022gram}              & 10.4 & 36.5& 0.955 & - & - & - & - & 7.67 \\
& GMPI \cite{zhao2022generative}        & 11.4 & \bf 39.8& \bf 0.977& 8.29  & \bf 39.0 & \bf 0.961 & -    & 7.79\\
& EpiGRAF \cite{skorokhodov2022epigraf} & 9.71  & - & - & 9.92  & 37.3 & 0.949 & 6.93 & - \\ 
& VoxGRAF \cite{schwarz2022voxgraf}     & 9.60  & 37.2 & 0.960 & -     & - & - & 9.60 & - \\
& Ours (3D branch)                                 & \bf 5.14  & 39.3 & 0.974 & \bf 5.37      & 37.8 & 0.955 & \bf 4.14  & \bf 4.29  \\

\hline
\end{tabular}
\caption{Comparison on image generation quality and 3D consistency among different 3D-aware GANs.}
\label{tab:render}
\end{table*}



\paragraph{Implementation details.}
We train our method on two real-world datasets: FFHQ~\cite{karras2019style} and AFHQ-v2 Cats~\cite{choi2020stargan}, which consists of 70K human face images of $1024^2$ resolution and 5.5K cat face images of $512^2$ resolution, respectively. We follow the data pre-processing of EG3D~\cite{chan2021efficient} to crop and resize the images to $256^2$ or $512^2$ resolution. Experiments are conducted on 8 NVIDIA Tesla A100 GPUs with
40GB memory, following the training configuration of EG3D. For FFHQ, the training process takes around 8 days, where learning the 2D branch takes 5 days and jointly training the whole framework takes additional 3 days. For AFHQ-v2, we finetune the 2D branch initially trained on FFHQ for 1 day, then jointly train the whole framework for extra 3 days. Adaptive data augmentation~\cite{karras2020training} is applied to AFHQ-v2 to facilitate training with limited data. 
See the \emph{suppl. material} for more details.


\subsection{Visual Results} \label{sec:visual}
Figure~\ref{fig:visual} shows the multiview images generated by our 3D branch generator. It can produce high-quality images with fine details at a resolution of $512^2$. Moreover, the images are of strict 3D consistency across different views via directly rendering the generated high-resolution NeRF. More results are in Fig.~\ref{fig:geo}, \ref{fig:compare}, and the \emph{suppl. material}.


\subsection{Comparison with Prior Arts}
\paragraph{Baselines.} We compare our method with existing 3D-aware GANs, including methods leveraging 2D super-resolution: StyleSDF~\cite{or2022stylesdf}, VolumeGAN~\cite{xu20223d}, StyleNeRF~\cite{gu2021stylenerf}, and EG3D~\cite{chan2021efficient}; and methods with direct 3D rendering: GRAM~\cite{deng2022gram}, GRAM-HD~\cite{xiang2022gram}, GMPI~\cite{zhao2022generative}, EpiGRAF~\cite{skorokhodov2022epigraf}, and VoxGRAF~\cite{schwarz2022voxgraf}.

\paragraph{Qualitative comparison.}
Figure~\ref{fig:geo} shows the visual comparison between our method and EG3D. Our generated images via direct rendering have comparable quality with those generated by EG3D via 2D super-resolution. We further visualize the 3D geometry and the spatiotemporal texture images~\cite{xiang2022gram} of the two methods. The geometry is extracted via Marching Cubes~\cite{lorensen1987marching} on the density field at $512^3$ resolution. The spatiotemporal textures are obtained by stacking the pixels of a fixed line segment under continuous camera change, very similar to the Epipolar Line Images~\cite{bolles1987epipolar}, where smoothly tilted strips indicate better 3D consistency. As shown, our geometries contain finer details in that we directly learn the NeRF of a subject at high resolution. Our spatiotemporal textures are also more reasonable with fewer twisted patterns, thanks to the direct 3D rendering for image synthesis instead of using a black-box 2D super-resolution module.

Figure~\ref{fig:compare} compares our method with other 3D baselines on FFHQ at $512^2$ resolution. Visually inspected, our 3D branch produces images of higher fidelity compared to existing methods leveraging direct 3D rendering. More analysis and video results can be found in the \emph{suppl. material}.

\vspace{-0.4cm}
\paragraph{Quantitative comparison.}
Table~\ref{tab:render} and Fig.~\ref{fig:teaser} show the quantitative results of different methods in terms of image generation quality and 3D consistency.
For image generation quality, We calculate the \'Frechet Inception Distance
(FID) \cite{heusel2017gans} between 50K generated images and all available real images in the training set. For 3D consistency, we follow GRAM-HD~\cite{xiang2022gram} to generate multiview images of 50 random subjects and train the multiview reconstruction method NeuS~\cite{wang2021neus} on each of them. We report the average PSNR and SSIM scores between our generated multiview images and the re-rendered images of NeuS (denoted as PSNR$_{mv}$ and SSIM$_{mv}$). Theoretically, better 3D consistency facilitates the 3D reconstruction process of NeuS, thus leading to higher PSNR and SSIM. 

As shown, our 2D branch generator demonstrates better results compared to EG3D in all metrics across different datasets, thanks to our 3D-aware stream in the tri-plane generator. Moreover, with the 3D-to-2D imitation strategy, our 3D branch generator largely improves the image generation quality among methods using direct 3D rendering, while maintaining competitive 3D consistency. Its image quality even surpasses most of the methods with 2D super-resolution and comes very close to that of EG3D.


\begin{table}[t]
\small
\renewcommand{\arraystretch}{1.2}
\centering
\begin{tabular}{c | c c c | c}
\hline
Label & $\mathcal{L}_{imitation}$ & $\mathcal S^{3D}$ & $\mathcal{L}_{adv}^{3D}$ & FID (3D branch) \\ 
\hline 
(A) &             &              &  & 30.6 \\
(B) & \checkmark  &              &  & 29.9 \\
(C) & \checkmark  & \checkmark   &  & 9.29 \\
(D) &             & \checkmark   & \checkmark & 22.8 \\
(E) & \checkmark  & \checkmark   & \checkmark &  \textbf{5.14}\\
\hline
\end{tabular}
\caption{Ablation study on 3D-to-2D imitation strategy.}
\label{tab:sr}
\vspace{-0.1cm}
\end{table}

\begin{figure}[t]
\begin{center}
\includegraphics[width=\linewidth]{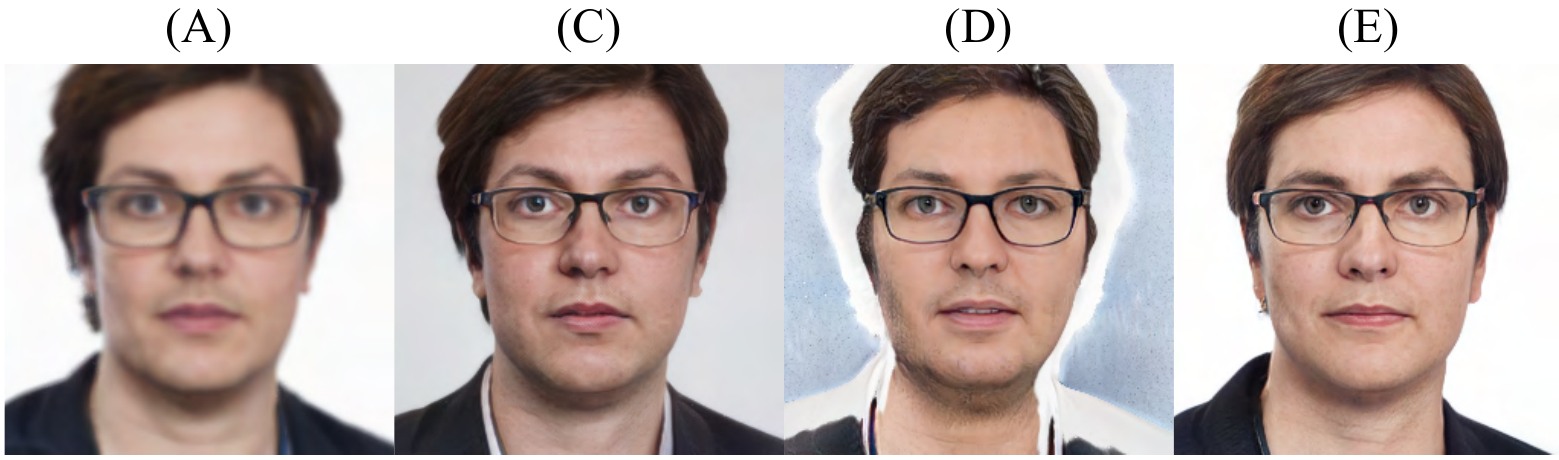}
\caption{Generated images under different learning strategies. The labels are consistent with Tab.~\ref{tab:sr}.}
\label{fig:sr}
\end{center}
\vspace{-0.5cm}
\end{figure}

\subsection{Ablation Study} \label{sec:ablation}
We conduct ablation studies to validate the efficacy of our proposed 3D-to-2D imitation and the 3D-aware tri-plane generator. For efficiency, all experiments are conducted on FFHQ dataset at $256^2$ resolution.

\vspace{-0.1cm}
\paragraph{3D-to-2D imitation strategy.} As shown in Tab.~\ref{tab:sr} and Fig.~\ref{fig:sr}, We start from a generator without using the 3D-to-2D imitation and the 3D super-resolution module $\mathcal{S}^{3D}$ (setting A), by directly rendering the coarse tri-planes $\mathbf{P}^c$ for image synthesis. The rendered images in this way are blurry and lack fine details, leading to a high FID score of $30.6$. Naively introducing the imitation loss (setting B) to improve the rendered images of $\mathbf{P}^c$ has minor influence, as the capacity of the coarse tri-planes are limited. Further incorporating the 3D super-resolution module (setting C) effectively releases the potential of the imitation loss and largely improves the image generation quality in terms of FID. However, the rendered images still lack rich details limited by the 3D-inconsistent 2D branch supervisions. Then, if the imitation loss is replaced with the adversarial loss (setting D), the image quality decreases significantly. This is due to that we only render small image patches to compute the corresponding losses for memory consideration. Under this circumstance, the adversarial loss is less stable compared to the imitation loss which is a perceptual-level reconstruction loss. This reveals the advantage of our imitation strategy, which could be extended to higher resolution via patch-wise optimization while maintaining a good image generation quality. Finally, leveraging all the three components (setting E) yields the best result, where the imitation loss keeps the overall structure reasonable and the adversarial loss helps with fine details learning.


\begin{table}[t]
\small
\setlength\tabcolsep{3pt}
\centering
\begin{tabular}{c | c c c c}
\hline
Method & FID (2D) & FID (3D)  & \#Param & Mem.  \\ 
\hline 
w/o 3D-aware   & 4.80 & 6.71 & 29.0M   & 2.3G   \\
\hline
3D-aware latent & OOM & OOM & 111.7M     & 11.6G \\
3D-aware tri-plane & 4.14 & - & 32.6M   & 2.4G   \\
3D-aware stream (Ours)  & \textbf{3.91} &\textbf{5.14}  & 32.6M   & 2.4G   \\
\hline
\end{tabular}
\caption{Ablation study on designs of 3D-aware tri-plane generator. The FID scores are from 2D or 3D branch; \#Param only considers the tri-plane generator $\mathcal{E}$ and Mem. indicates the GPU memory cost for generating the coarse tri-planes.}
\label{tab:aware3d}
\vspace{-0.1cm}
\end{table}

\begin{figure}[t]
\begin{center}
\includegraphics[width=\linewidth]{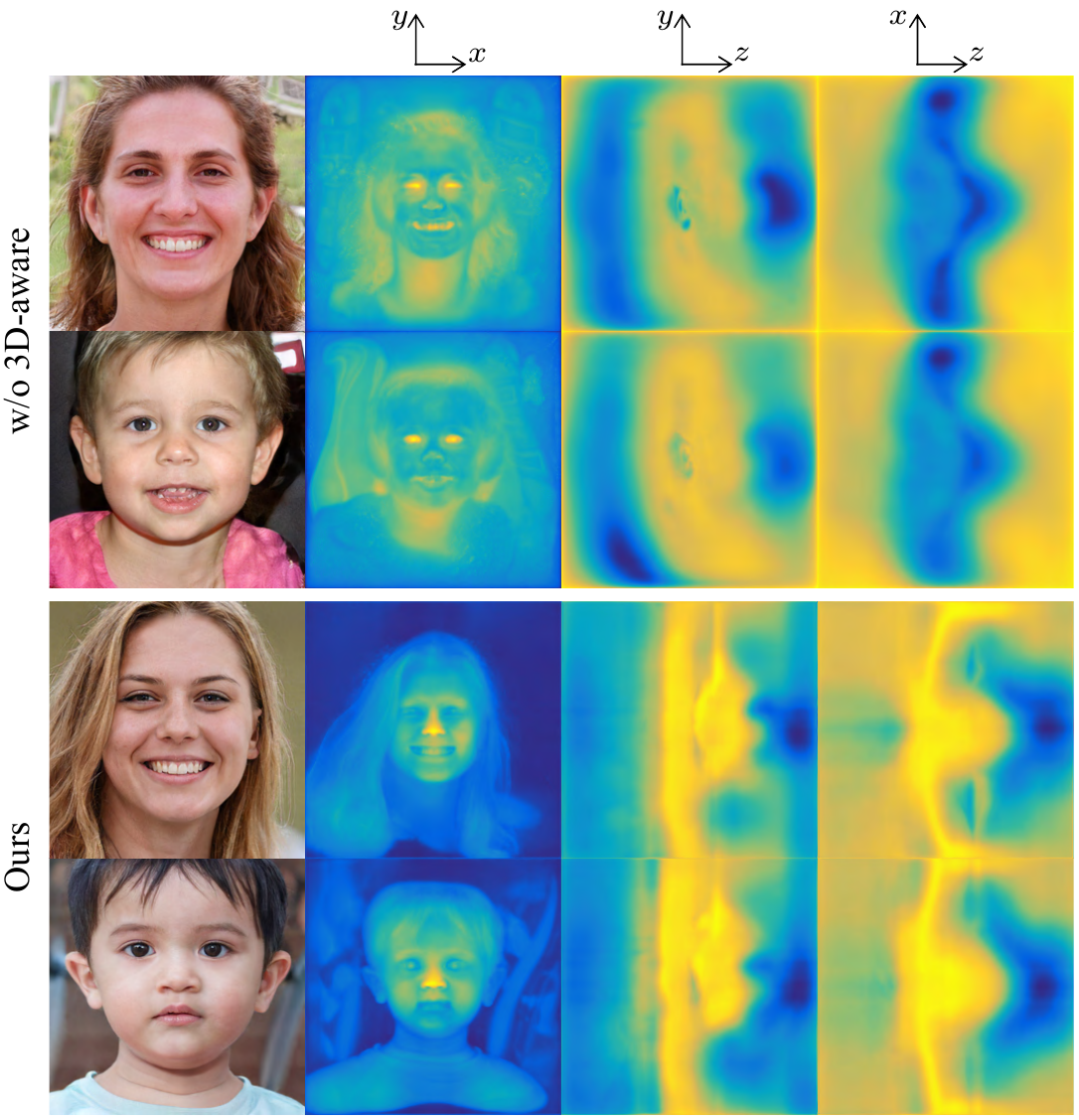}
\caption{Generated tri-planes with or w/o 3D-aware convolutions.}
\label{fig:plane}
\end{center}
\vspace{-0.5cm}
\end{figure}

\vspace{-0.3cm}
\paragraph{3D-aware tri-plane generator.} Table~\ref{tab:aware3d} shows the ablation study on the 3D-aware tri-plane generator. We compare our design with two alternatives and one without 3D-aware convolutions originally adopted by EG3D. We report the parameter size of the tri-plane generators, the inference memory cost to generate the coarse tri-planes, as well as the final image generation quality in terms of FID. In the first alternative, we remove our 3D-aware stream, and leverage 3D-aware convolutions for the latent feature maps in the main stream, namely \textit{3D-aware latent}. Since the main stream feature maps have relatively larger feature channels, and the 3D-aware convolution requires to concatenate two additional tensors with the same size as the input tensor, this design increases the parameter size and memory consumption significantly, and raises the out-of-memory issue during training. In the second alternative, namely \textit{3D-aware tri-plane}, we directly apply 3D-aware convolutions in the output stream, by inserting them after the upsampling operations at each resolution, instead of using the additional 3D-aware stream. This strategy leads to an improvement of the image generation quality of the 2D branch, and largely reduces the parameter size and memory cost compare to the first design. Finally, our 3D-aware stream design further improves the image generation quality without introducing extra parameters and memory costs. Therefore, we adopt it as our final 3D-aware tri-plane generator for 3D-to-2D imitation. It effectively lowers the FID score of both the 2D and 3D branches compared to the original structure without 3D-aware convolutions, with only a minor increase of the parameter size and memory cost.

Figure~\ref{fig:plane} further shows the synthesized tri-planes, where we visualize the L2 norm of each spatial location on the three orthogonal planes. Our method leveraging the 3D-aware stream produces more informative tri-planes. The generated planes of the side-views better depict the characters of different instances (\eg, see the difference of the profiles on the $yz$ planes). Our frontal planes (\ie $xy$ planes) also demonstrate more clear head silhouettes compared to those without using the 3D-aware convolutions.



\section{Conclusions}
We presented a novel learning strategy for 3D-aware GANs to achieve image synthesis of high-quality and strict 3D consistency. The core idea is to enforce the images synthesized by the generator's 3D rendering branch to mimic those generated by its 2D super-resolution branch. We also introduced 3D-aware convolutions to the generator to further improve the image generation quality. With the above strategies, our method largely improves the image quality among methods using direct 3D rendering, which we believe enables a new way for more realistic 3D generation. 

\vspace{-0.3cm}
\paragraph{Limitation and future works.} Our method has several limitations. The image generation quality of its 3D branch still lags behind that of the 2D branch. Certain generated 3D structures such as hairs and cat whiskers are stuck to the geometry surfaces instead of correctly floating in the volumetric space. The 3D-to-2D imitation strategy also introduces extra training time and memory costs compared to only learning the 2D branch. We expect more effective learning strategies and more advanced 3D representations to alleviate these problems.

\vspace{-0.3cm}
\paragraph{Ethics consideration.} The goal of this paper is to generate images of virtual subjects. It is not intended for creating misleading or deceptive contents of real people and we do not condone any such harmful behavior.

{\small

}

\clearpage
\appendix

\section{More Implementation Details}

\subsection{Network Structure}
Figure~\ref{fig:net} illustrates our network designs, including the 3D super-resolution module $\mathcal S^{3D}$ and the 3D-aware block in the tri-plane generator $\mathcal{E}$.


For $\mathcal S^{3D}$ (Fig.~\ref{fig:net}(a)), we use two modulated 2D convolution blocks~\cite{karras2020analyzing} to upsample the tri-planes.

For the 3D-aware block (Fig.~\ref{fig:net}(b)), we re-organize the tri-planes according to Fig. 4 in the main text, and apply modulated 2D convolutions for each of the three planes. We use different affine layers to generate style codes for the three modulated convolutions, respectively.

\subsection{Training Details}
We randomly sample latent code $z$ from the normal distribution and camera pose $\boldsymbol\theta$ from those of the training datasets to synthesize fake images, following EG3D~\cite{chan2021efficient}. For each viewing ray, we sample 96 points to calculate the volume rendering equation, including 48 points with stratified sampling and 48 points with importance sampling. The learning rates of the generator and the two discriminators are set to 0.0025 and 0.002, respectively. We train the 2D branch with 25M images in total, and then jointly train the whole framework with additional 15M images. The batch size during training is set as 32.
Other training settings are identical to those of EG3D \cite{chan2021efficient}. 

\subsection{Patch Scale}
To reduce GPU memory costs and enable training at high resolution, we render $64^2$ patches for the 3D-to-2D imitation. Thus, the patch scale is $1/4$ or $1/8$ of the whole image for the $256^2$ or $512^2$ experiments, respectively. The patch center is uniformly sampled from the whole image space.

\subsection{The necessity of 2D super-resolution module}
The function of the 2D super-resolution in the 2D branch is to provide stable and high-quality guidance for the 3D branch. 
Previous studies have attempted to directly learn in 3D space without 2D super-resolution via the adversarial loss. However, due to the restriction of modern GPU memory, they either adopted more efficient 3D representations (\eg, radiance manifolds~\cite{deng2022gram} or MPI~\cite{zhao2022generative}) or used patch-wise loss (\eg, EpiGRAF \cite{skorokhodov2022epigraf}), yet these strategies often lead to worse diversity and image quality due to the instability of the GAN loss. By contrast, our imitation with the 2D branch via LPIPS loss provides stable gradients for learning the 3D representation, and thus supports patch-wise training without sacrificing the generation quality, which is the key to our superior results. Furthermore, our strategy also avoids troublesome training tricks (\eg, the annealed strategy in EpiGRAF \cite{skorokhodov2022epigraf}) thus easier to be adapted to other frameworks. 

\subsection{Training time/memory of 3D-to-2D imitation}
Our method requires 31 GB memory at $256^2$ resolution with a batch size of 32 when trained on 8 GPUs, compared to 27 GB memory without the 3D-to-2D imitation. Also, our training time is 1.5 times longer than that of EG3D.

\section{More Results and Comparisons}

\begin{figure}[t]
\begin{center}
\includegraphics[width=\linewidth]{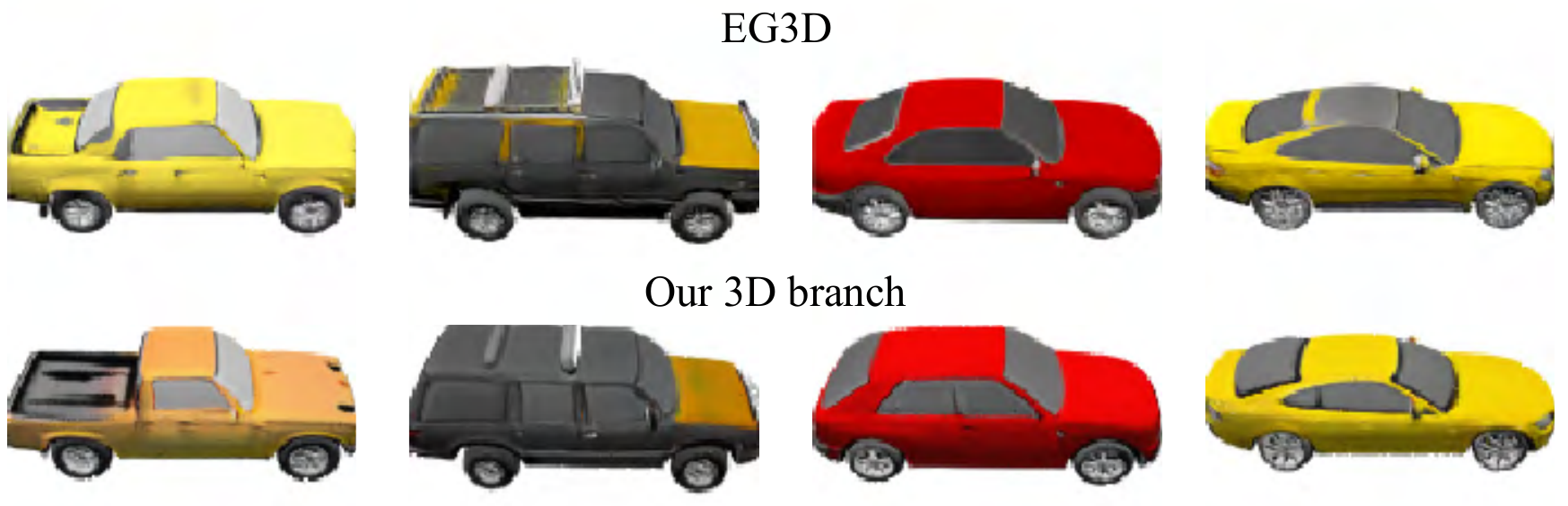}
\caption{Comparison with EG3D on ShapeNet-Cars.}
\label{fig:car}
\end{center}
\vspace{-0.3cm}
\end{figure}

\begin{figure}[t]
\begin{center}
\includegraphics[width=\linewidth]{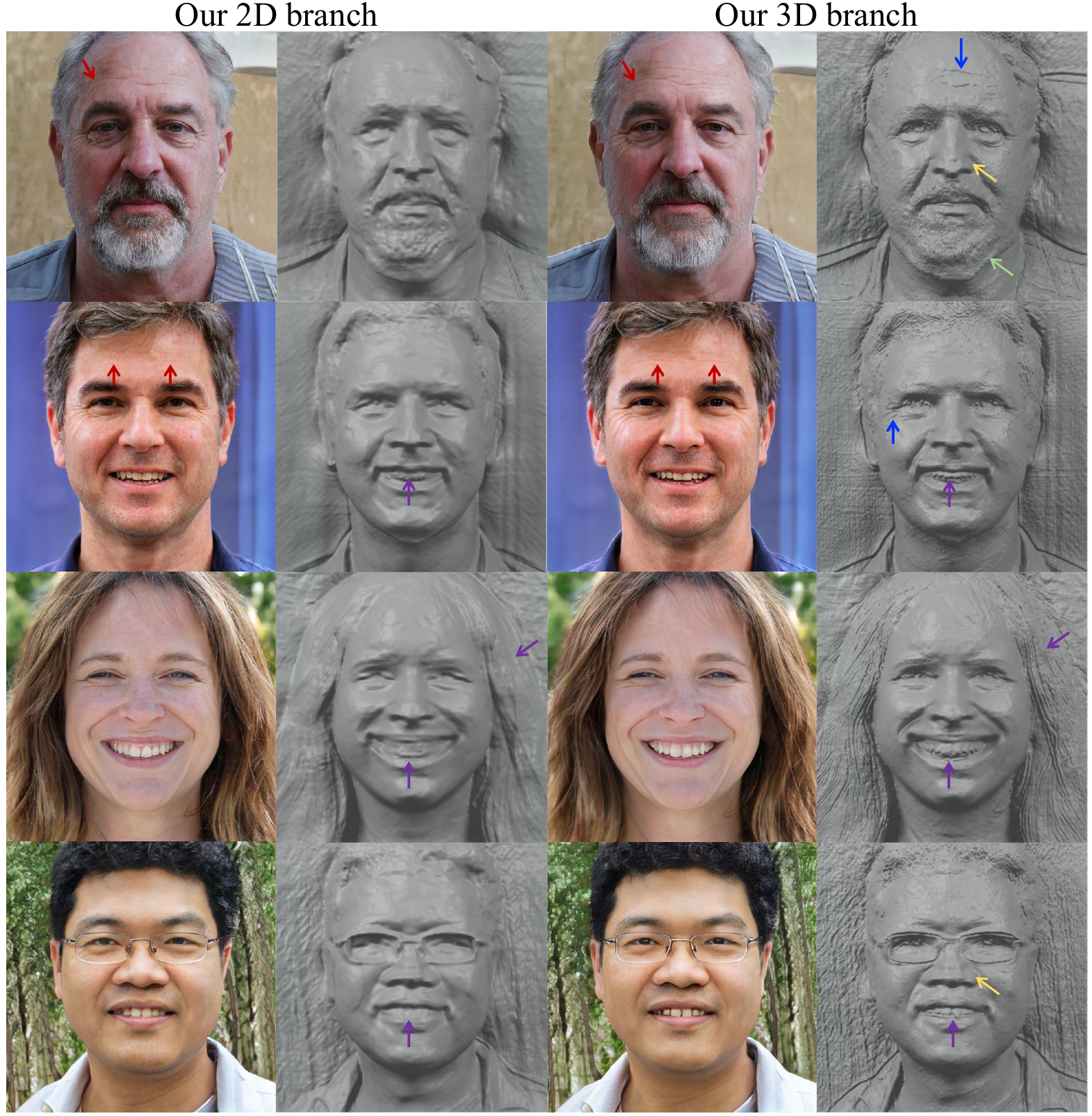}
\caption{Comparison of 2D and 3D branches. (Zoom in for better visualization.)}
\label{fig:moregeo}
\end{center}
\end{figure}

\subsection{End-to-end 3D-to-2D imitation learning}
Our initial motivation for the two-stage training is to leverage the powerful prior of an existing 2D generator (with 2D super-resolution) to guide our 3D branch. In fact, the overall framework (including both 2D and 3D branches) can be trained end-to-end from scratch. We conduct a simple experiment on FFHQ at $256^2$ with identical hyper parameters as described in the main paper and achieve an FID of $5.03$ for the 3D branch, which is comparable to the two-stage training result.

\subsection{More results on faces}
Figures~\ref{fig:mtconst} and \ref{fig:mtcomp} illustrate more visual comparisons. Compared to EG3D \cite{chan2021efficient}, we have more detailed geometry and smoothly tilted strips in spatiotemporal texture images, indicating better 3D consistency. Similar to ours, EpiGRAF and GMPI also generate high-resolution images via direct rendering. Yet, we have superior image quality as shown in Fig.~\ref{fig:mtcomp}.

Figures~\ref{fig:mtface} and \ref{fig:mtcat} show more of our results on FFHQ and AFHQ -v2 Cats datasets, respectively.

\textit{\textbf{Referring to the supplemental video for animations.}}

\subsection{Results on general objects.}
Our method can handle general objects with wider range of camera views. In Fig.~\ref{fig:car}, we compare our 3D branch with EG3D on ShapeNet-Cars ($128^2$) and achieve comparable image generation quality.

\subsection{Comparison of our 2D and 3D branches}

Our 3D branch can generate fine details comparable to the 2D branch. In Fig.~\ref{fig:moregeo} (red arrows), we show details produced by the 3D branch that are not visible in the 2D branch.

Our 3D branch clearly produces finer geometry details compared to the alternatives with 2D super-resolution (see Fig.~\ref{fig:moregeo}). As shown, the finer geometry details are not random noises but features of hair, teeth, wrinkles, etc (purple arrows). Furthermore, we can generate diverse nose shapes (yellow arrows), complex jaws with beards (green arrows), and wrinkles (blue arrows) on the geometries.

\begin{figure}[t]
\begin{center}
\includegraphics[width=\linewidth]{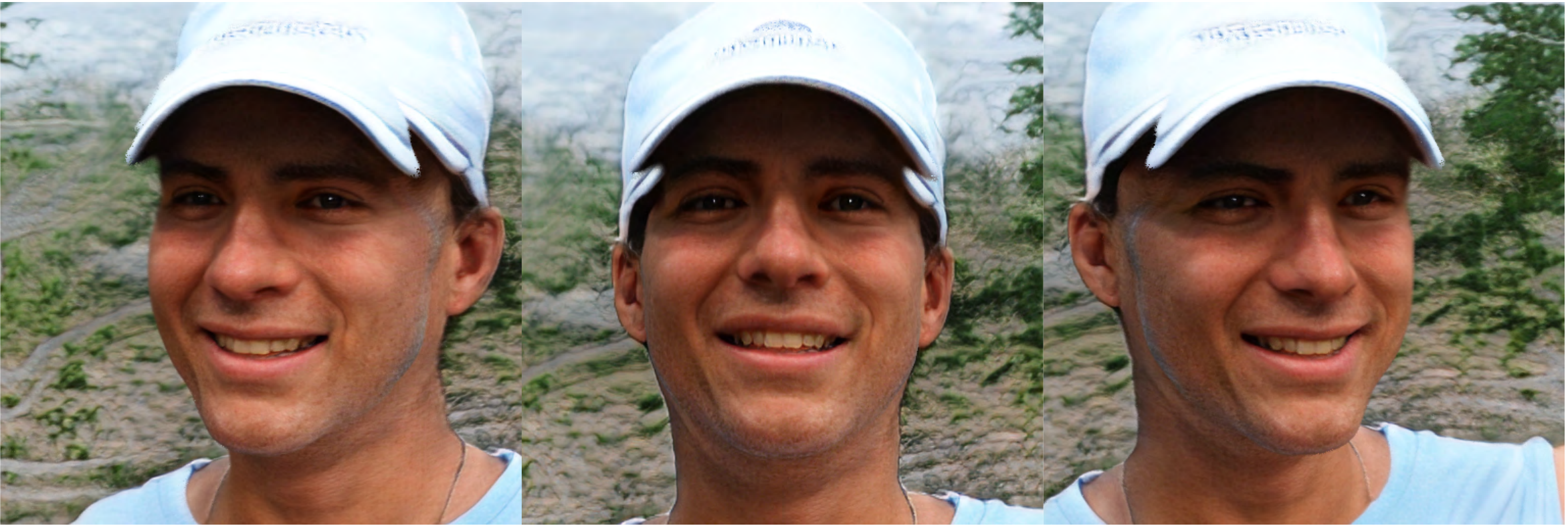}
\caption{Failure case.}
\label{fig:fail}
\end{center}
\vspace{-0.3cm}
\end{figure}

\section{Limitations and Future Works}

We thoroughly discuss the limitations of our method and possible future improvements. 

First, our learned 3D branch still has inferior image quality in terms of FID compared to the 2D branch. This may come from the current design of the 3D super-resolution module and the learning strategy. Specifically, our 3D super-resolution module adopts a similar structure to that of the 2D branch in order for a fair comparison, which may not be the optimal solution. More advanced structures, including leveraging 3D-aware convolutions could be further explored for better 3D super-resolution. Besides, the LPIPS loss during 3D-to-2D imitation leverages a pre-trained VGG network which is trained on images of $224^2$ resolution. It may not well capture the perceptual information of a small image patch. Leveraging more recent pre-trained models~\cite{he2020momentum,radford2021learning} or even multiple feature extractors could be a possible choice. Exploring better discriminators for the patch-level adversarial loss in the 3D branch could also benefit the training process.

Second, our method can produce incorrect geometries in certain cases. As shown in Fig.~\ref{fig:fail}, a typical failure case is geometry discontinuity, where the face region is not smoothly connected with the head region, leading to obvious artifacts at side views. These artifacts also occur in the original EG3D. We believe this problem can be alleviated by introducing more profile images for training, as currently the training data are mostly frontal images so that the planes for depicting side-view features may not be well-trained. In addition, certain generated geometry structures such as hairs and cat whiskers are stuck to the surfaces instead of correctly floating in the volumetric space, as shown in Fig.~\ref{fig:mtface} and~\ref{fig:mtcat}. We conjecture this is due to that the random sampling strategy with limited queries during volume rendering is hard to model thin structures, as also indicated by a previous method~\cite{deng2022gram}. Therefore, a more advanced 3D representation that could efficiently capture these complex structures is worthy of ongoing exploration.

Finally, our training strategy also requires training the 2D branch in advance, which increases the overall training time compared to the original EG3D. A possible way to reduce the training time is to jointly train the 2D and 3D branches from scratch. We leave it for our future work.





\begin{figure*}[t]
\begin{center}
\includegraphics[width=0.85\linewidth]{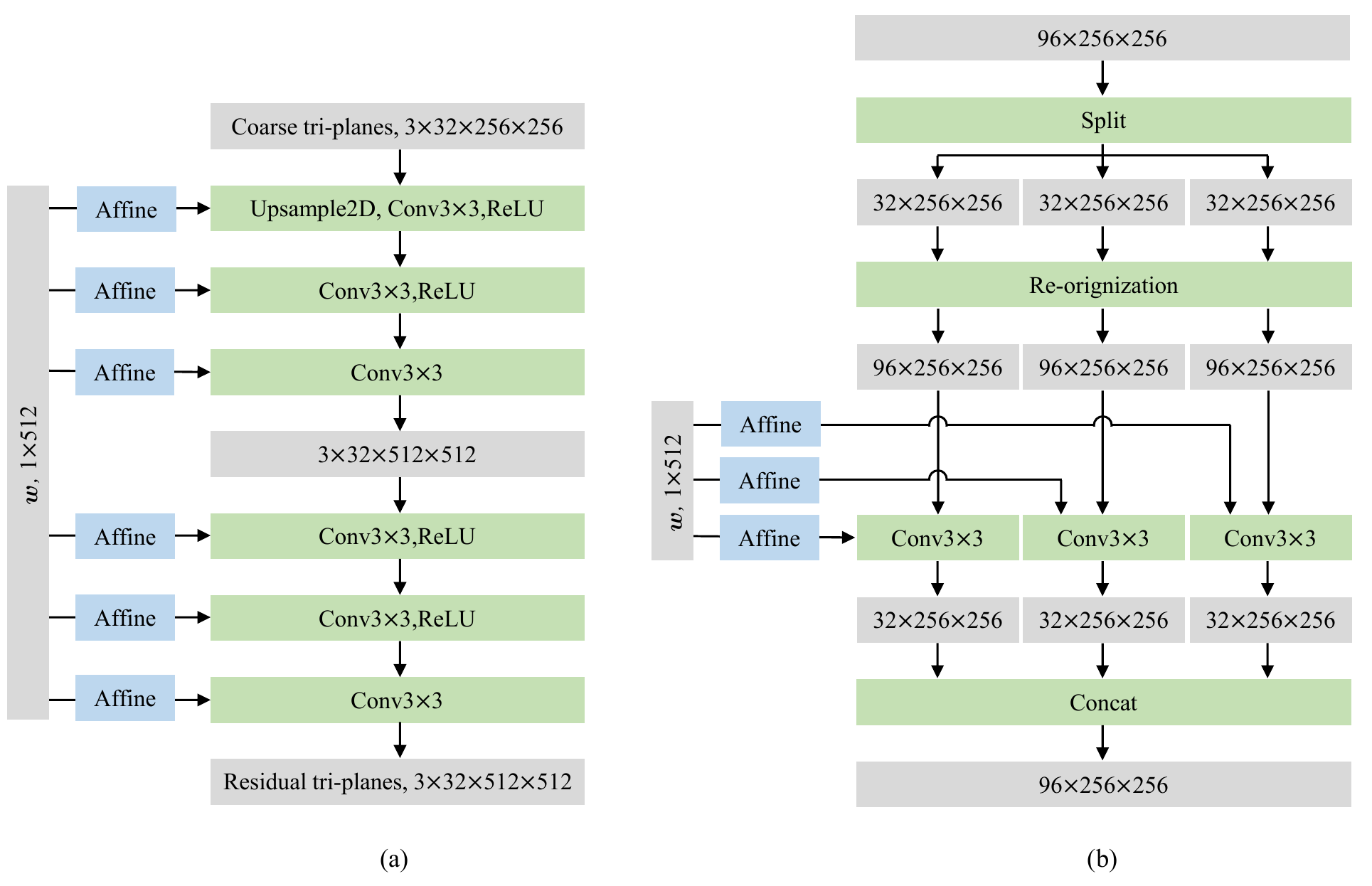}
\caption{Network designs. (a) 3D super-resolution module $\mathcal S^{3D}$. (b) 3D-aware block.}
\label{fig:net}
\end{center}
\vspace{-0.4cm}
\end{figure*}

\begin{figure*}[b]
\begin{center}
\includegraphics[width=\linewidth]{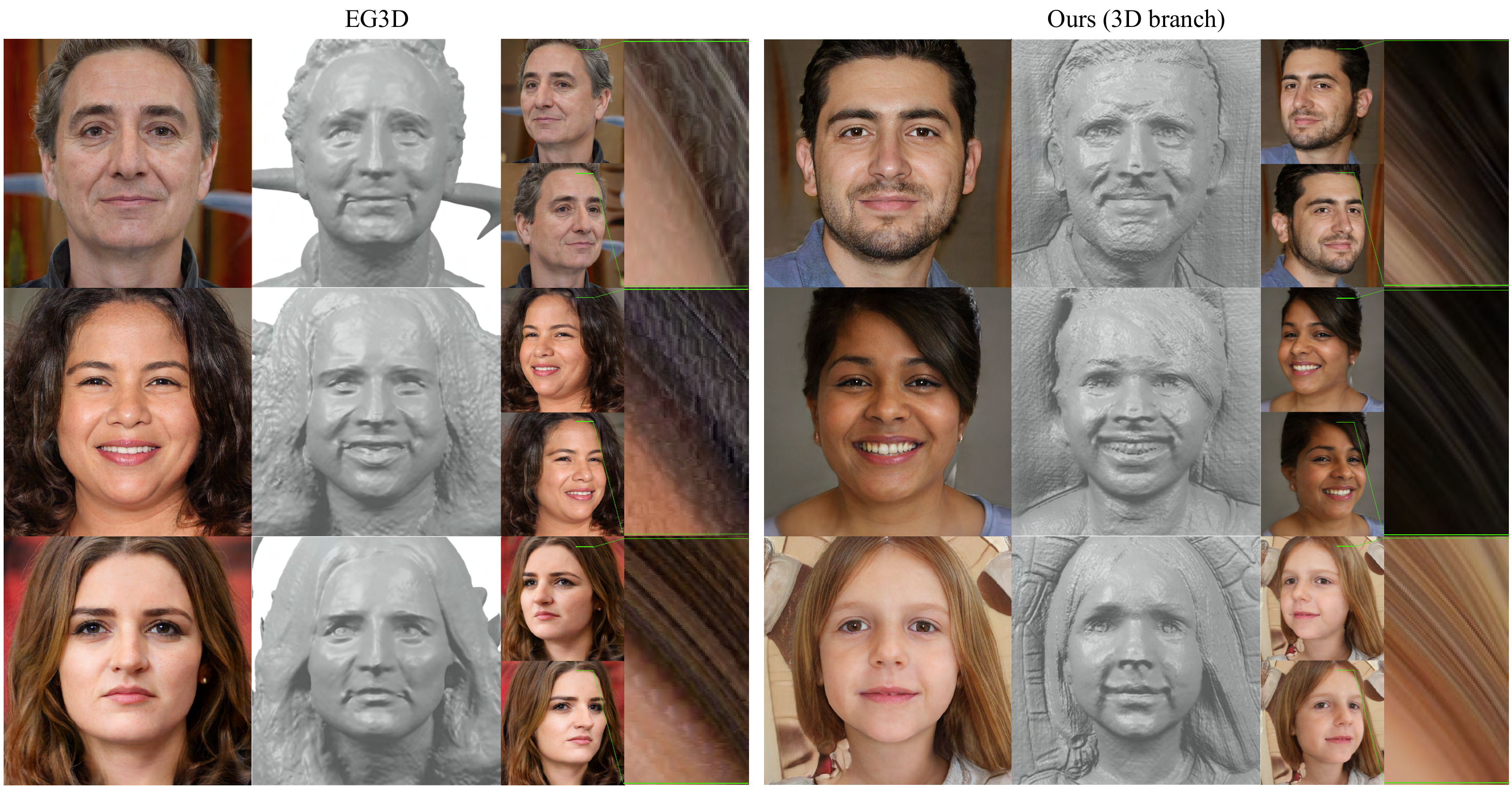}
\caption{Comparison w/ EG3D \cite{chan2021efficient}. Our method generates images with comparable quality to those of EG3D, while producing 3D geometries with finer details and multiview sequences with better 3D-consistency. Referring to the supplemental video for animations.}
\label{fig:mtconst}
\end{center}
\vspace{-0.4cm}
\end{figure*}

\begin{figure*}[t]
\begin{center}
\includegraphics[width=\linewidth]{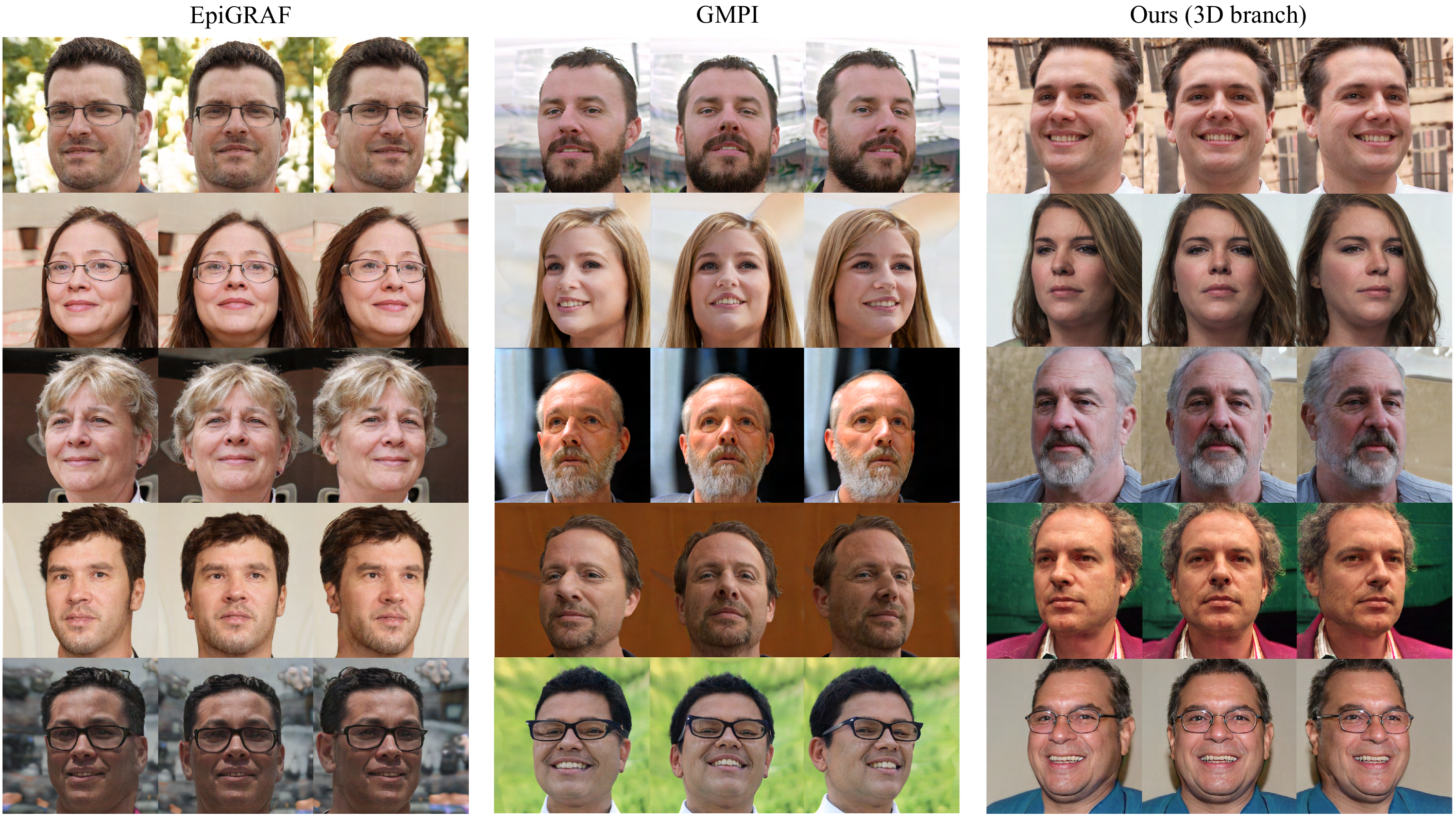}
\caption{Comparison w/ EpiGRAF \cite{skorokhodov2022epigraf} and GMPI \cite{zhao2022generative} . Referring to the supplemental video for animations.}
\label{fig:mtcomp}
\end{center}
\vspace{-0.4cm}
\end{figure*}

\begin{figure*}[t]
\begin{center}
\includegraphics[width=\linewidth]{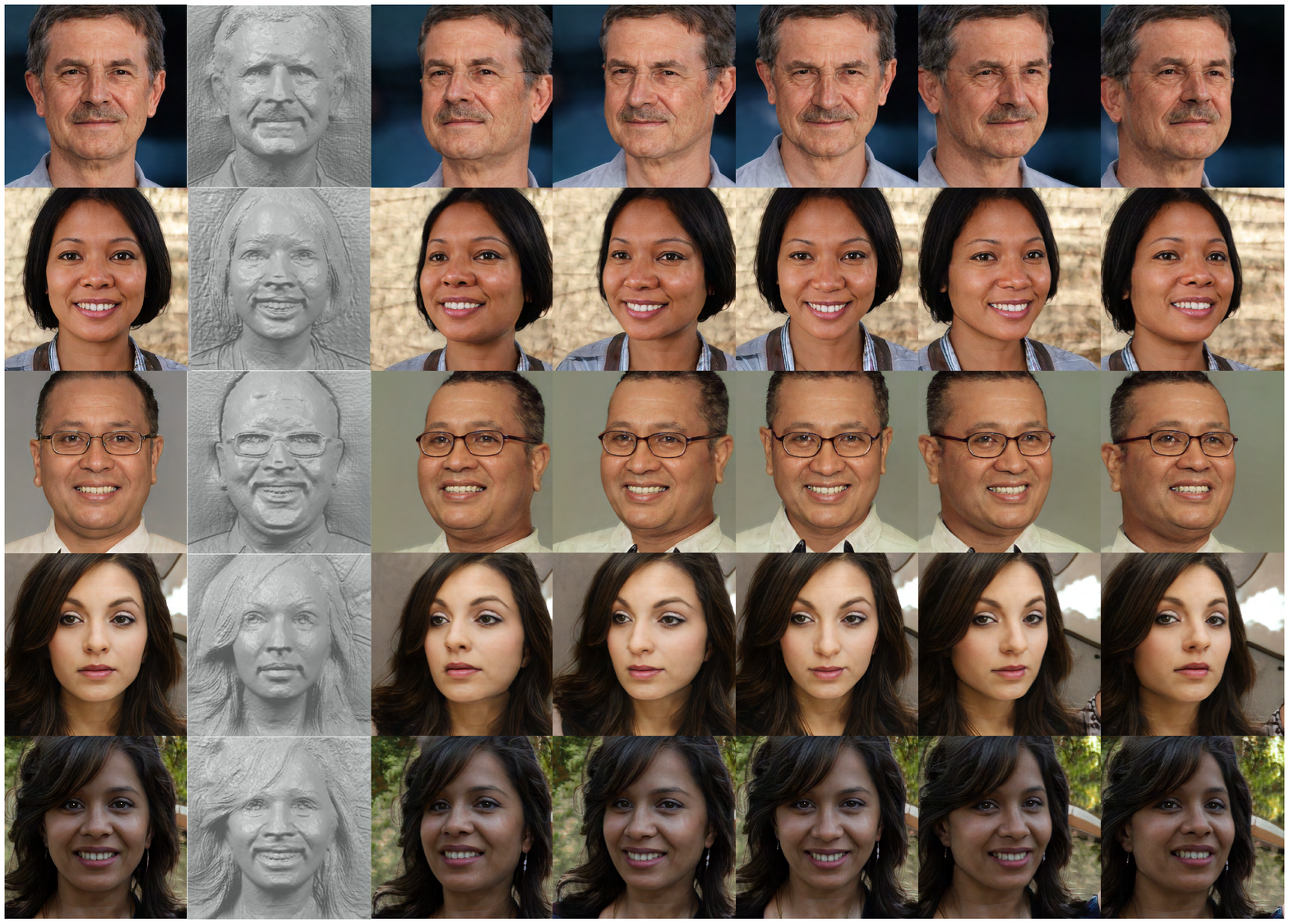}
\caption{Our results on FFHQ dataset. Referring to the supplemental video for animations.}
\label{fig:mtface}
\end{center}
\vspace{-0.4cm}
\end{figure*}

\begin{figure*}[t]
\begin{center}
\includegraphics[width=\linewidth]{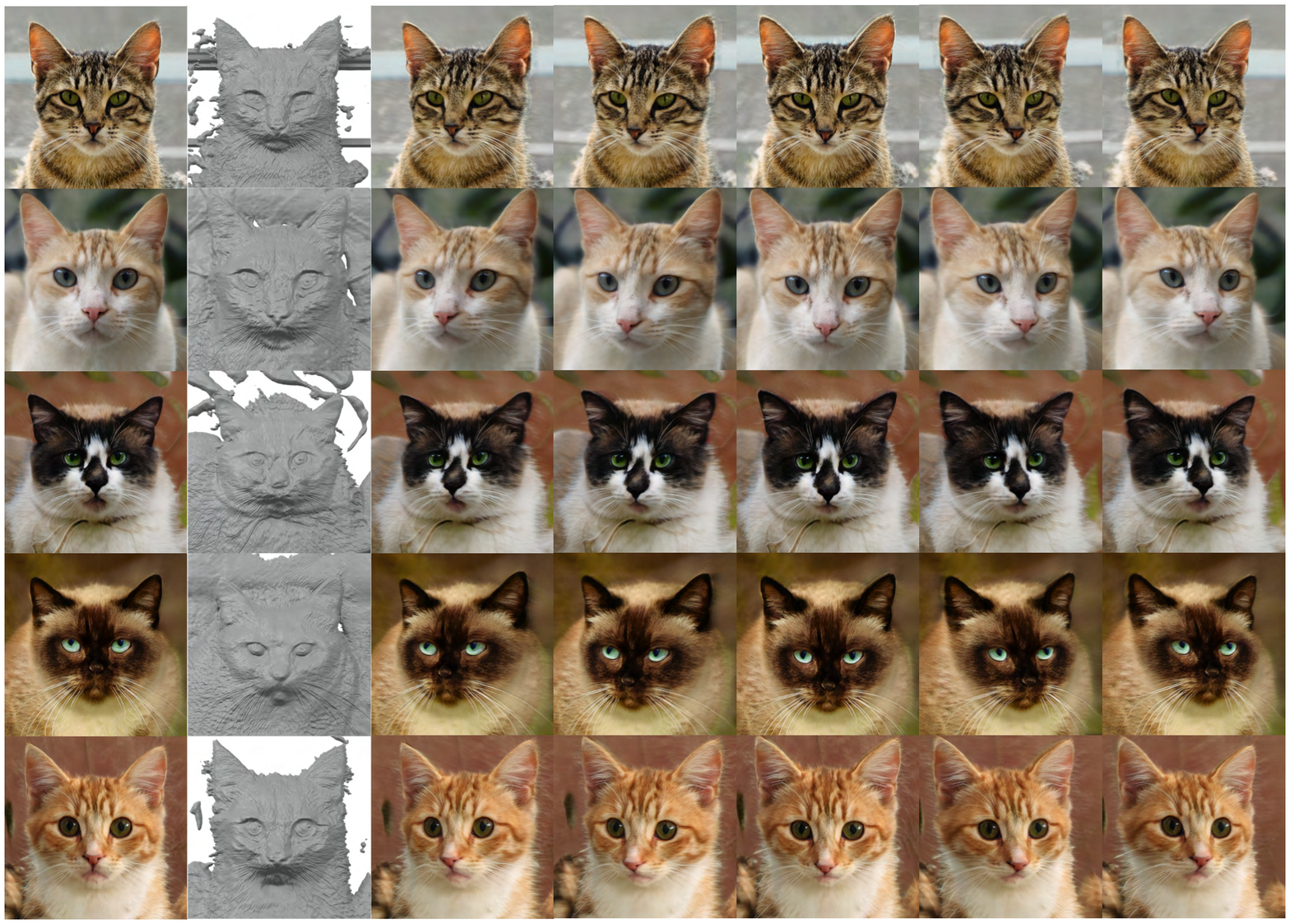}
\caption{Our results on AFHQ-v2 Cats. Referring to the supplemental video for animations.}
\label{fig:mtcat}
\end{center}
\vspace{-0.4cm}
\end{figure*}

\end{document}